\documentclass[a4paper, 10pt, conference]{ieeeconf}      
\usepackage{FG2021}

\FGfinalcopy 

\usepackage{graphicx}
\usepackage{amsmath}
\usepackage{amssymb}

\usepackage{algorithm}
\usepackage{algorithmic}
\usepackage{multirow}
\usepackage{array}

\IEEEoverridecommandlockouts                              
\def\FGPaperID{204} 

\title{\LARGE \bf
Federated Test-Time Adaptive Face Presentation Attack Detection with Dual-Phase Privacy Preservation
}

\author{\parbox{16cm}{\centering
		{\normalsize Rui Shao$^1$ \qquad Bochao Zhang$^1$ \qquad Pong C. Yuen$^1$ \qquad Vishal M. Patel$^2$}\\
		{\normalsize
			$^1$ Department of Computer Science, Hong Kong Baptist University, Hong Kong\\
			$^2$ Department of Electrical and Computer Engineering, Johns Hopkins University, USA}}
}

\begin{document}

\ifFGfinal
\thispagestyle{empty}
\pagestyle{empty}
\else


\pagestyle{plain}
\fi
\maketitle

\begin{abstract}
Face presentation attack detection (fPAD) plays a critical role in the modern face recognition pipeline. The generalization ability of face presentation attack detection models to unseen attacks has become a key issue for real-world deployment, which can be improved when models are trained with face images from different input distributions and different types of spoof attacks. In reality, due to legal and privacy issues, training data (both real face images and spoof images) are not allowed to be directly shared between different data sources. In this paper, to circumvent this challenge, we propose a Federated Test-Time Adaptive Face Presentation Attack Detection with Dual-Phase Privacy Preservation framework, with the aim of enhancing the generalization ability of fPAD models in both training and testing phase while preserving data privacy. In the training phase, the proposed framework exploits the federated learning technique, which simultaneously takes advantage of rich fPAD information available at different data sources by aggregating model updates from them without accessing their private data. To further boost the generalization ability, in the testing phase, we explore test-time adaptation by minimizing the entropy of fPAD model prediction on the testing data, which alleviates the domain gap between training and testing data and thus reduces the generalization error of a fPAD model. We introduce the experimental setting to evaluate the proposed framework and carry out extensive experiments to provide various insights about the proposed method for fPAD.

\end{abstract}

\section{Introduction}

Recent advances in face recognition methods have prompted many real-world applications, such as automated teller machines (ATMs), mobile devices, and entrance guard systems, to deploy this technique as an authentication method. Wide usage of this technology is due to both high accuracy and convenience it provides. However,  many recent works~\cite{2011IJCBmstexture,2016TIFScolortxt,2015TIFSida,RuiShao2018IJCB,2018CVPRauxliary,Shao2019CVPR,2018TIFSdynamictext,Shao_2020_AAAI} have found that this technique is vulnerable to various face presentation attacks such as print attacks, video-replay attacks~\cite{2017FGoulu,2012ICBcasia,2012BIOSIGidiap,2015TIFSida,2018CVPRauxliary} and 3D mask attacks~\cite{2018ECCVrPPG,2016ECCVrPPG}. Therefore, developing face presentation attack detection (fPAD) methods that make current face recognition systems robust to face presentation attacks has become a topic of interest in the biometrics community.

In this paper, we consider the deployment of a fPAD system in the real-world scenario. We identify two types of stakeholders in this scenario -- \textit{data centers} and \textit{users}. \textit{Data centers} are entities that design and collect fPAD datasets and propose fPAD solutions. Typically \textit{data centers} include research institutions and companies that carry out the research and development of fPAD. These entities have access to both \textit{real data} and \textit{spoof data} and therefore are able to train fPAD models. Different \textit{data centers} may contain images of different identities and different types of \textit{spoof data}. However, each \textit{data center} has  limited data availability. Real face images are obtained from a small set of identities and spoof attacks are likely to be from  a few known types of attacks. Therefore, these fPAD models have poor generalization ability~\cite{Shao2019CVPR,Shao_2020_AAAI} and are likely to be vulnerable against attacks unseen during training.

\begin{figure}[!t]
	
	\begin{center}
		
		\includegraphics[ width=0.65\linewidth]{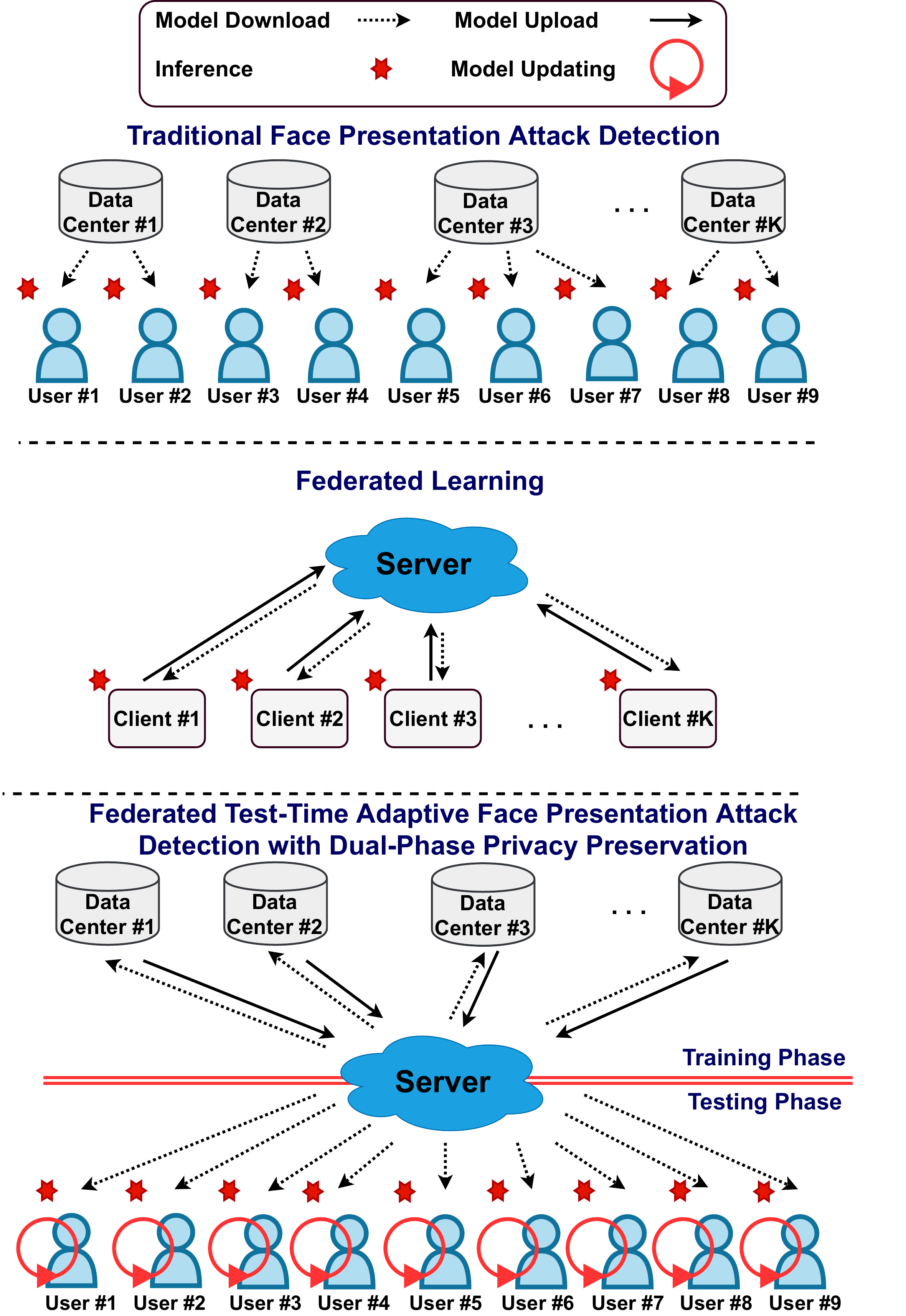}
		
	\end{center}
	
	\vskip-12pt\caption{Comparison between fPAD (top), traditional federated learning (middle) and the proposed framework (bottom).}
	\label{fig:illustration}
\end{figure}

On the other hand, \textit{users} are individuals or entities that make use of fPAD solutions. For example, when a fPAD algorithm is introduced in mobile devices, mobile device customers are identified as \textit{users} of the fPAD system.  \textit{Users} have access only to \textit{real data} collected from local devices. Due to the absence of \textit{spoof data}, they cannot locally train fPAD models. Therefore, each \textit{user} relies on a model developed by a \textit{data center} for fPAD as shown in Figure~\ref{fig:illustration} (top). Since \textit{data center} models lack generalization ability, inferencing with these models are likely to result in erroneous predictions. 

It has been shown that utilizing \textit{real data} from different input distributions and \textit{spoof data} from different types of spoof attacks through domain generalization and meta-learning techniques can significantly improve the generalization ability of fPAD models~\cite{Shao2019CVPR,Shao_2020_AAAI}. Therefore, the performance of fPAD models, shown in Figure~\ref{fig:illustration} (top), can be improved if data from all \textit{data centers} can be exploited collaboratively. In reality, due to data sharing agreements and privacy policies, \textit{data centers} are not allowed to share collected fPAD data with each other. For example, when a \textit{data center} collects face images from individuals using a social media platform, it is agreed not to share collected data with third parties. 

In this paper, we present a framework called Federated Test-Time Adaptive Face Presentation Attack Detection with Dual-Phase Privacy Preservation based on the principles of Federated Learning (FL) and test-time adaptation targeting fPAD. Federate learning is a distributed and privacy preserving machine learning technique~\cite{mcmahan2016communication,li2019federated,smith2017federated,sahu2018convergence,mohri2019agnostic}. FL training paradigm defines two types of roles named \textit{server} and \textit{client}.  \textit{Clients} contain training data and the capacity to train a model.  As shown in Fig.~\ref{fig:illustration} (middle), each client trains its own model locally and uploads them to the \textit{server} at the end of each training iteration. \textit{Server} aggregates local updates and produces a global model. This global model is then shared with all clients which will be used in their subsequent training iteration. This process is continued until the global model is converged. During the training process, data of each client is kept private. Collaborative FL training allows the global model to exploit rich local clients information while preserving data privacy.   

In the context of FL for fPAD, both \textit{data centers} and \textit{users} can be identified as clients. However, roles of \textit{data centers}  and \textit{users} are different from conventional clients found in FL. In FL, all \textit{clients} train models and carry out inference locally. In contrast, in FL for fPAD, only \textit{data centers} carry out local model training. \textit{Data centers} share their models with the \textit{server} and download the global model during the training phase. On the other hand, \textit{users} download the global model at the end of the training procedure and carry out inference in the testing phase as shown in Figure~\ref{fig:illustration} (bottom).

Although domain gaps between training and testing data can be alleviated by the above federated learning process carried out in the training phase, fPAD models are still hard to generalize well to unseen attacks with significant domain variations in the real-world deployment. To further improve the generalization ability of fPAD model, we incorporate test-time adaptation into our framework by exploiting hints provided by the testing data. Specifically, such hints can be unveiled from the entropy of model predictions on the testing data and generalization error of fPAD models can be reduced via minimization of entropy of model predictions. As illustrated in Figure~\ref{fig:illustration} (bottom), after downloading the trained fPAD models from the training phase, \textit{users} further adapt the fPAD model in the testing phase with the help of entropy minimization of model predictions before the final classification. Note that test-time adaptation is sensitive to the initial parameters of a pre-trained model. Federated learning can produce a well pre-trained model with privacy preservation which provides a good starting point for the following test-time adaptation. Two phases are compatible with each other and together improve the generalization ability of fPAD without accessing private training data from multiple data centers.

%
%

\begin{figure}[!htb]
	
	\begin{center}
		
		\includegraphics[width=\linewidth]{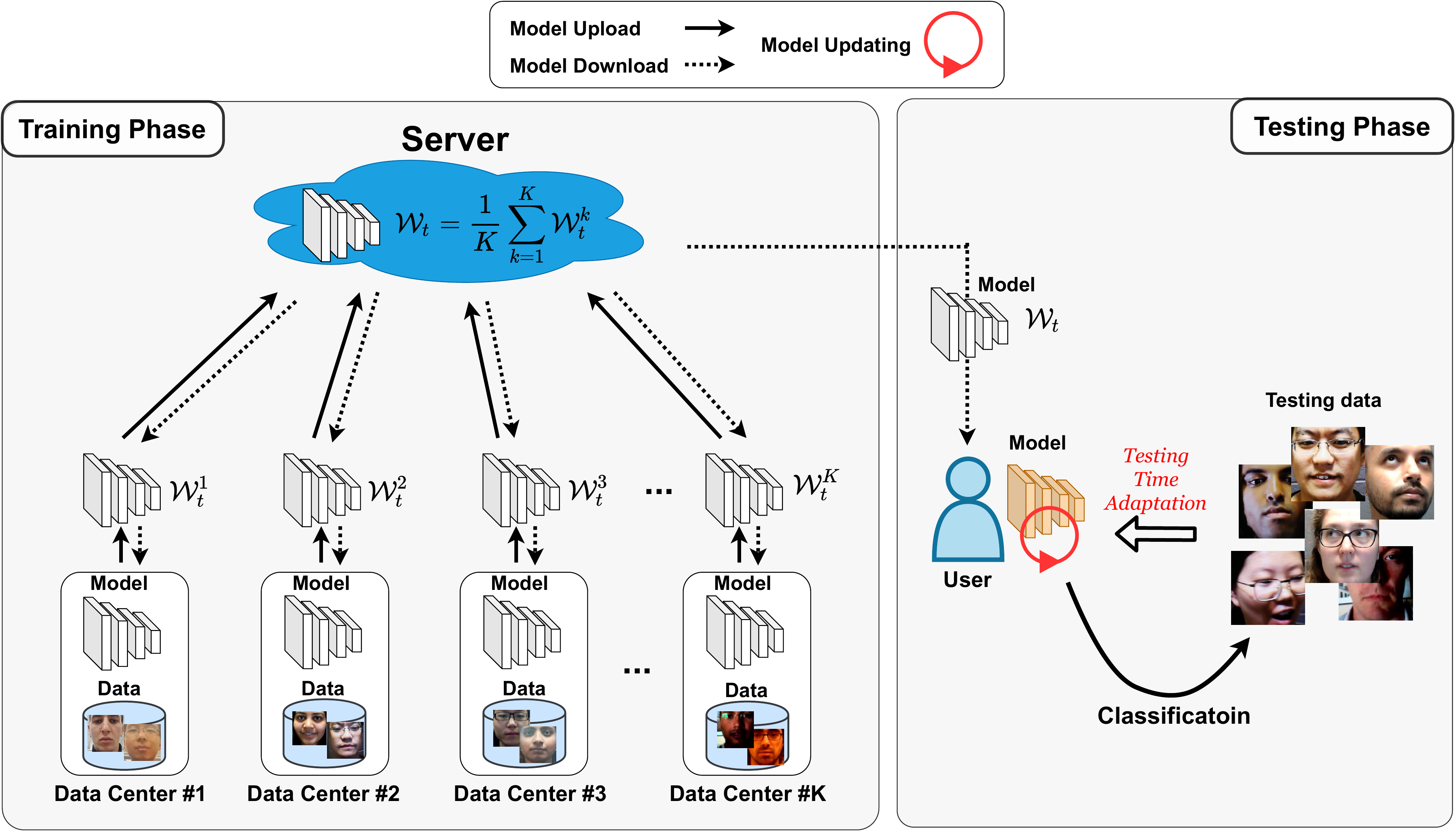}
		
	\end{center}
	
	\vskip-15pt\caption{Overview of the proposed framework. In the training phase, through several rounds of communications between \textit{data centers} and \textit{server}, the collaborated trained global fPAD model parameterized by $\mathcal{W}_t$ can be obtained in a data privacy preserving way. In the testing phase, users download this global model from the server to their device and further carry out testing time adaptation given the testing data at hand. The final classification for the testing data can be carried out based on the updated fPAD model.}
	\label{fig:Overview}
\end{figure}

\section{Related Work}

\subsection{Face Presentation Attack Detection}
Current fPAD methods can be categorized under single-domain and multi-domain approaches. Single-domain approaches focus on extracting discriminative cues between real and spoof samples from a single dataset, which can be further divided into appearance-based methods and temporal-based methods. Appearance-based methods focus on extracting various discriminative appearance cues for detecting face presentation attacks. Multi-scale LBP~\cite{2011IJCBmstexture} and color textures~\cite{2016TIFScolortxt} methods are two texture-based methods that extract various LBP descriptors in various color spaces for the differentiation between real/spoof samples. Image distortion analysis~\cite{2015TIFSida} aims to detect the surface distortions as the discriminative cue.  

On the other hand, temporal-based methods extract different discriminative temporal cues through multiple frames between real and spoof samples. Various dynamic textures are exploited in~\cite{2014EJIVPlbptop,2018TIFSdynamictext,RuiShao2018IJCB} to extract discriminative facial motions. rPPG signals are exploited by Liu \emph{et al.}~\cite{2016ECCVrPPG,2018ECCVrPPG} to capture discriminative heartbeat information from real and spoof videos. ~\cite{2018CVPRauxliary} learns a CNN-RNN model to estimate different face depth and rPPG signals between the real and spoof samples. 

Various fPAD datasets are introduced recently that explore different characteristics and scenarios of face presentation attacks. Multi-domain approach is proposed in order to improve the generalization ability of the fPAD model to unseen attacks. Recent work~\cite{Shao2019CVPR} casts fPAD as a domain generalization problem and proposes a multi-adversarial discriminative deep domain generalization framework to search generalized differentiation cues in a shared and discriminative feature space among multiple fPAD datasets. ~\cite{liu2019deep} treats fPAD as a zero-shot problem and proposes a Deep Tree Network to partition the spoof samples into multiple sub-groups of attacks. ~\cite{Shao_2020_AAAI} addresses fPAD with a meta-learning framework and enables the model learn to generalize well through simulated train-test splits among multiple datasets. These multi-domain approaches have access to data from multiple datasets or multiple spoof sub-groups that enable them to obtain generalized models. In this paper, we study the scenario in which each \textit{data center} contains data from a single domain. Due to data privacy issues, we assume that they do not have access to data from other \textit{data centers}. This paper aims to exploit  multi-domain information in a privacy preserving manner.

\subsection{Federated Learning}
Federated learning is a decentralized machine learning approach that enables multiple local clients to collaboratively learn a global model with the help of a server while preserving data privacy of local clients. Federated averaging (FedAvg)~\cite{mcmahan2016communication}, one of the fundamental frameworks for FL, learns a global model by averaging model parameters from local clients. FedProx~\cite{sahu2018convergence} and Agnostic Federated Learning (AFL)~\cite{mohri2019agnostic} are two variants of FedAvg which aim to address the bias issue of the learned global model towards different clients. These two methods achieve better models  by adding proximal term to the cost functions and optimizing a centralized distribution mixed with client distributions, respectively.  A recent work FedPAD~\cite{shao2020federated} also exploits federated learning for  the task of fPAD. However, when encountering unseen attacks with large domain gap, the performance is still degraded. Comparatively, the proposed method integrates the testing-time adaptation with federated learning, which further facilities the generalization ability of a fPAD model during testing.

\section{Proposed Method}

\subsection{Training Time Federated Learning}

The proposed Test-Time Adaptive Face Presentation Attack Detection framework is summarized in Fig.~\ref{fig:Overview} and Algorithm~\ref{algorithm}. Suppose that $K$ \textit{data enters} collect their own fPAD datasets designed for different characteristics and scenarios of face presentation attacks. The corresponding collected fPAD datasets are denoted as $\mathcal{D}^{1}, \mathcal{D}^{2},..., \mathcal{D}^{K}$ with data provided with image and label pairs denoted as $x$ and $y$. $y$ denotes the ground-truth with binary class labels (y= 0/1 are the labels of spoof/real samples). Based on the collected fPAD data, each \textit{data center} can train its own fPAD model by iteratively minimizing the cross-entropy loss as follows:
\setlength{\belowdisplayskip}{0pt} \setlength{\belowdisplayshortskip}{0pt}
\setlength{\abovedisplayskip}{0pt} \setlength{\abovedisplayshortskip}{0pt}

\begin{equation}
\nonumber 		\mathcal{L}(\mathcal{W}^k) = \sum\limits_{(x,y)\sim\mathcal{\mathcal{D}}^{k}}y\log\mathcal{F}^k(x)+(1-y)\log(1-\mathcal{F}^k(x)),
\end{equation} 
where the fPAD model $\mathcal{F}^{k}$ of the $k$-th \textit{data enter} is parameterized by $\mathcal{W}^k$ ($k=1,2,3,...,K$). After optimization with several local epochs via $$\mathcal{W}^k \leftarrow \mathcal{W}^k-\eta \nabla\mathcal{L}(\mathcal{W}^k),$$ each \textit{data enter} can obtain the trained fPAD model with the updated model parameters.

It should be noted that dataset corresponding to each  \textit{data center} is from a specific input distribution and it only contains  a finite set of known types of spoof attack data. When a  model is trained using this data, it focuses on addressing the characteristics and scenarios of face presentation attacks prevalent in the corresponding dataset. However,  a model trained from a specific \textit{data center} will not generalize well to unseen face presentation attacks. It is a well known fact that diverse fPAD training data contributes to a better generalized fPAD model. A straightforward solution is to collect and combine all the data from $K$ data centers denoted as $\mathcal{D} = \{\mathcal{D}^{1} \cup \mathcal{D}^{2}\cup...\cup \mathcal{D}^{K}\} $ to train a fPAD model. It has been shown that domain generalization and meta-learning based fPAD methods can further improve the generalization ability with the above combined multi-domain data $\mathcal{D}$~\cite{Shao2019CVPR,Shao_2020_AAAI}. However, when sharing data between different \textit{data centers} are prohibited due to the privacy issue, this naive solution is not practical.

\begin{algorithm}[t]
	\small
	\caption{Federated Test-Time Adaptive Face Presentation Attack Detection with Dual-Phase Privacy Preservation}
	\begin{algorithmic}
		\REQUIRE~~\\
		\textbf{Input:} $K$ Data Centers have $K$ fPAD datasets $\mathcal{D}^{1}, \mathcal{D}^{2},..., \mathcal{D}^{K}$,  Testing data presented to user $\mathcal U$\\
		\textbf{Initialization:} $K$ Data Centers have $K$ fPAD models $\mathcal{F}^{1}, \mathcal{F}^{2},..., \mathcal{F}^{K}$ parameterized by $\mathcal{W}_0^1, \mathcal{W}_0^2,..., \mathcal{W}_0^K$. $L$ is the number of local epochs. $\eta$ is the learning rate. $t$ is the federated learning rounds
		
		\STATE \textbf{Training Phase:}
		
		\textbf{Server aggregates:}
		
		initialize $\mathcal{W}_0$
		\FOR{each round $t$ = 0, 1, 2,... }
		\FOR{each data center $k$ = 1, 2,..., $K$  \textbf{in parallel}}
		\STATE $  {W}_t^k \leftarrow {\rm \textbf{DataCenterUpdate}}(k, \mathcal{W}_t)$
		\ENDFOR
		\STATE $\mathcal{W}_t=\frac{1}{K} \sum\limits_{k=1}^K\mathcal{W}_t^k$
		\STATE 	\textbf{Download} ${W}_t$ to \textbf{Data Centers}
		\ENDFOR
		\STATE \textbf{Testing Phase:}
		\STATE \textbf{Users} \textbf{Download} model $\mathcal{F}_t$ parameterized by $\mathcal{W}_t$
		\STATE $\mathcal{H}(\mathcal{F}_t(x)) = \sum\limits_{x\sim\mathcal{\mathcal{U}}}\mathcal{F}_t(x)\log\mathcal{F}_t(x)+(1-\mathcal{F}_t(x))\log(1-\mathcal{F}_t(x)) $
		\STATE $\mathcal{W}_{t(\gamma,\beta)} \leftarrow \mathcal{W}_{t(\gamma,\beta)}-\eta \nabla\mathcal{H}(\mathcal{F}_t(x))$
		\STATE\textbf{Users make final classification}
		
		\STATE
		\STATE
		${\rm \textbf{DataCenterUpdate}}(k, \mathcal{W})$\textbf{:}
		\FOR{each local epoch $i$ = 1, 2,..., $L$  }
		\STATE $\mathcal{L}(\mathcal{W}^k) = \sum\limits_{(x,y)\sim\mathcal{\mathcal{D}}^{k}}y\log\mathcal{F}^k(x)+(1-y)\log(1-\mathcal{F}^k(x))$
		\STATE $\mathcal{W}^k \leftarrow \mathcal{W}^k-\eta \nabla\mathcal{L}(\mathcal{W}^k)$
		\ENDFOR
		
		\textbf{Upload} ${W}^k$ to \textbf{Server}
	\end{algorithmic}
	\label{algorithm}
\end{algorithm}

To circumvent this limitation and enable various data centers to collaboratively train a fPAD model, we propose the Test-Time Adaptive Face Presentation Attack Detection framework. Instead of accessing private fPAD data of each \textit{data center}, the proposed framework introduces a \textit{server} to exploit the fPAD information of all data centers by aggregating the above model updates ($\mathcal{W}^1, \mathcal{W}^2,..., \mathcal{W}^K$) of all data centers. Inspired by the  Federated Averaging~\cite{mcmahan2016communication} algorithm, in the proposed framework, server carries out the aggregation of model updates via calculating the average of updated parameters ($\mathcal{W}^1, \mathcal{W}^2,..., \mathcal{W}^K$) in all data centers as $\mathcal{W}=\frac{1}{K} \sum\limits_{k=1}^K\mathcal{W}^k$.

After the aggregation, server  produces a global fPAD model parameterized by $\mathcal{W}$ that exploits the fPAD information of various data centers without accessing the private fPAD data. We can further extend the above aggregation process into $t$ rounds. Server distributes the  aggregated model $\mathcal{W}$ to every data center as the initial model parameters for the next-round updating of local parameters. Thus, data centers can obtain the $t$-th round updated parameters denoted as ($\mathcal{W}_t^1, \mathcal{W}_t^2,..., \mathcal{W}_t^K$). The $t$-th aggregation in the server can be carried out as $	\mathcal{W}_t=\frac{1}{K} \sum\limits_{k=1}^K\mathcal{W}_t^k$.
After  $t$-rounds of communication between data centers and the server, the trained global fPAD model parameterized by $\mathcal{W}_t$ can be obtained without compromising the private data of individual \textit{data centers}. Once training is converged, \textit{users} will download the trained model from the server to their devices to carry out fPAD locally.

\subsection{Testing-Time Adaptation}

\begin{figure}[!t]
	
	\begin{center}
		
		\includegraphics[width=1\linewidth]{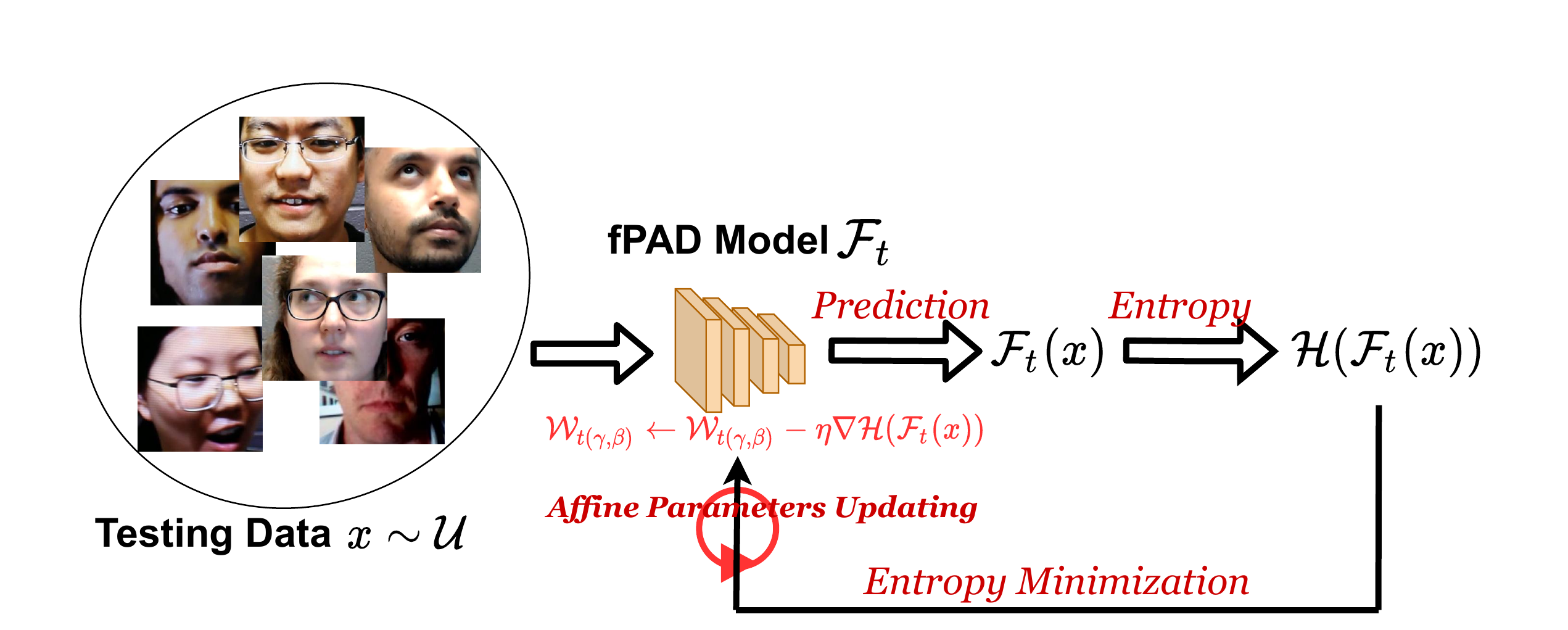}
		
	\end{center}
	
	\vskip-15pt\caption{Details of test-time adaptation during testing. Note that test-time adaptation updates the parameters of a fPAD model using unlabeled testing data presented to the users without accessing the training data.}
	\label{fig:TTT}
\end{figure}

Although federated learning in the training phase can exploit various fPAD information available from multiple data centers to improve the generalization ability of a fPAD model, the fPAD model can not generalize well to some unseen attacks when the domain gap between the training and testing data is large, i.e., significant environmental variations. To address this issue, we further propose the test-time adaptation. We argue that the data presented to the users during testing can provide some hints about their distribution, which can be used to adapt the fPAD model before making the final classification. Specifically, as demonstrated in~\cite{wang2021tent}, such hints can be unveiled by Shannon entropy~\cite{shannon1948mathematical} of model predictions which can indicate the distribution shift between seen training data and unseen testing data in an unsupervised way. Therefore, the domain gap between training and testing data can be further reduced by minimizing the Shannon entropy of the fPAD model predictions on the testing data. As illustrated in Fig.~\ref{fig:TTT}, given the testing data presented to user $\mathcal{U}$, and the fPAD model $\mathcal{F}_t$ downloaded from the training phase, we calculate the entropy of fPAD model prediction as follows:

\begin{equation}
\nonumber 	\mathcal{H}(\mathcal{F}_t(x)) = \sum\limits_{x\sim\mathcal{\mathcal{U}}}\mathcal{F}_t(x)\log\mathcal{F}_t(x)+(1-\mathcal{F}_t(x))\log(1-\mathcal{F}_t(x)).
\end{equation}

To reduce the probability of overfitting during test-time adaptation given limited testing data, we focus on minimizing the above entropy with respect to affine transformation parameters (scale $\mathcal{W}_{t(\gamma)}$ and shift $\mathcal{W}_{t(\beta)}$) of all batch normalization layers in the fPAD model and keep all the other parameters fixed. This process can be carried out as: $\mathcal{W}_{t(\gamma,\beta)} \leftarrow \mathcal{W}_{t(\gamma,\beta)}-\eta \nabla\mathcal{H}(\mathcal{F}_t(x))$. After this test time adaptation, we use the updated fPAD model for the final real/fake classification. Same as federated learning carried out in the training phase, the whole process of the above test-time adaptation also does not need to get access to the private training data from multiple data centers, and thus the generalization ability of fPAD model can be further ameliorated in a privacy preserving manner during testing. Moreover, the above entropy calculation does not require one to design a specific self-supervision task such as image rotation for the supervision of adaptation, which significantly facilitates its compatibility to the task of fPAD. 

On the other hand, we should note that test-time adaptation is sensitive to the quality of initial parameters of a pre-trained model. Federated learning can produce a well pre-trained model in a privacy preserving manner so that a suitable starting point can be provided for the following optimization of testing-time adaptation. This further demonstrates the compatibility of dual-phase privacy preservation for the task of fPAD.

\section{Experiments}
\begin{figure*}[!htb]
	\begin{center}
		\includegraphics[height=2.5cm, width=.9\linewidth]{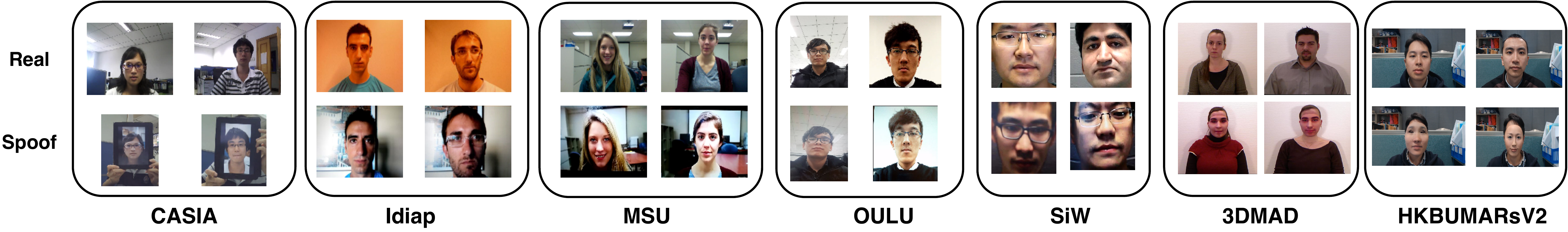}
	\end{center}
	\vskip-15pt\caption{Sample images corresponding to real and attacked faces from CASIA-MFSD~\cite{2012ICBcasia}, Idiap Replay-Attack~\cite{2012BIOSIGidiap}, MSU-MFSD~\cite{2015TIFSida}, Oulu-NPU~\cite{2017FGoulu}, SiW~\cite{2018CVPRauxliary}, 3DMAD~\cite{Marceltifs3D2014}, and HKBUMARsV2~\cite{2018ECCVrPPG} datasets.}
	\label{fig:datasets}
\end{figure*}

To evaluate the performance of the proposed framework, we carry out extensive experiments using five 2D fPAD datasets and two 3D mask fPAD datasets. In this section, we first describe the datasets and the testing protocol used in our experiments. Then we report various experimental results based on multiple fPAD datasets. Discussions and analysis about the results are carried out to provide various insights about FL for fPAD.

\subsection{Experimental Settings}

\subsubsection{Datasets}

\begin{table}[htb]
	\renewcommand{\arraystretch}{1}
	\centering	
	\scriptsize
	\caption{Comparison of seven experimental datasets.}	
	\begin{tabular}{c|c|c|c|c}
		\hline\hline
		\textbf{Dataset} & \begin{tabular}[c]{@{}c@{}}\textbf{Extra }\\\textbf{ light}\end{tabular} & \begin{tabular}[c]{@{}c@{}}\textbf{Complex}\\ \textbf{background}\end{tabular} & \begin{tabular}[c]{@{}c@{}}\textbf{Attack}\\ \textbf{type}\end{tabular}                                  & \begin{tabular}[c]{@{}c@{}}\textbf{Display} \\ \textbf{devices}\end{tabular}           \\ \hline
		C       & No                                                     & Yes                                                           & \begin{tabular}[c]{@{}c@{}}Printed photo\\ Cut photo\\ Replayed video\end{tabular}     & iPad                                                                 \\ \hline
		I       & Yes                                                    & Yes                                                          & \begin{tabular}[c]{@{}c@{}}Printed photo\\ Display photo\\ Replayed video\end{tabular} & \begin{tabular}[c]{@{}c@{}}iPhone 3GS \\ iPad\end{tabular}           \\ \hline
		M       & No                                                     & Yes                                                          & \begin{tabular}[c]{@{}c@{}}Printed photo\\ Replayed video\end{tabular}                 & \begin{tabular}[c]{@{}c@{}}iPad Air\\ iPhone 5S\end{tabular}         \\ \hline
		O       & Yes                                                    & No                                                           & \begin{tabular}[c]{@{}c@{}}Printed photo\\ Display photo\\ Replayed video\end{tabular} & \begin{tabular}[c]{@{}c@{}}Dell 1905FP\\ Macbook Retina\end{tabular} \\ \hline
		
		S       & Yes                                                    & Yes	                                                           & \begin{tabular}[c]{@{}c@{}}Printed photo\\ Display photo\\ Replayed video\end{tabular} & \begin{tabular}[c]{@{}c@{}}Dell 1905FP\\iPad Pro\\iPhone 7\\Galaxy S8\\ Asus MB168B\end{tabular} \\ \hline
		
		3       & No                                                    & No	                                                           & \begin{tabular}[c]{@{}c@{}}Thatsmyface 3D mask\end{tabular} & \begin{tabular}[c]{@{}c@{}}Kinect \end{tabular} \\ \hline
		H       & Yes                                                    & Yes	                                                           & \begin{tabular}[c]{@{}c@{}}Thatsmyface 3D mask\\REAL-f mask\end{tabular} & \begin{tabular}[c]{@{}c@{}}MV-U3B\end{tabular} \\ \hline\hline
	\end{tabular}
	\label{tab:datasets}
\end{table}

We evaluate our method using the following seven publicly available fPAD datasets which contain print, video replay and 3D mask attacks:\\
1) Oulu-NPU~\cite{2017FGoulu} (O for short)\\
2) CASIA-MFSD~\cite{2012ICBcasia} (C for short)\\ 
3) Idiap Replay-Attack~\cite{2012BIOSIGidiap} (I for short)\\ 
4) MSU-MFSD~\cite{2015TIFSida} (M for short)\\
5) SiW~\cite{2018CVPRauxliary} (S for short)\\
6) 3DMAD~\cite{Marceltifs3D2014} (3 for short)\\
7) HKBUMARsV2~\cite{2018ECCVrPPG} (H for short).\\ 
Table~\ref{tab:datasets} shows the variations in these seven datasets. Some sample images from these datasets are shown in Fig.~\ref{fig:datasets}. From Table~\ref{tab:datasets} and Fig.~\ref{fig:datasets} it can be seen that different fPAD datasets exploit different characteristics and scenarios of face presentation attacks (\textit{i.e.} different attack types, display materials and resolution, illumination, background and so on). Therefore, significant  domain shifts exist among these datasets.  

\subsubsection{Protocol} 
The testing protocol used in the paper is designed to test the generalization ability of fPAD models. Therefore, in each test, performance of a trained model is evaluated against a dataset that it has not been observed during training. In particular, we choose one dataset at a time to emulate the role of \textit{users} and consider all other datasets as \textit{data centers}. Real images and spoof images of \textit{data centers} are used to train a fPAD model. The trained model is tested considering the dataset that emulates the role of \textit{users}. We evaluate the performance of the model by considering how well the model is able to differentiate between real and spoof images belonging to each \textit{user}.

\subsubsection{Evaluation Metrics} 
Half Total Error Rate (HTER)~\cite{2004HTER} (half of the summation of false acceptance rate and false rejection rate), Equal Error Rates (EER) and Area Under Curve (AUC) are used as evaluation metrics in our experiments, which are three most widely-used metrics for the cross-datasets/cross-domain evaluations. Following ~\cite{liu2019deep}, in the absence of a development set,  thresholds required for calculating evaluation metrics are determined based on the data in all \textit{data centers}.

\subsubsection{Implementation Details} 
Our deep network is implemented on the platform of PyTorch. We adopt Resnet-18~\cite{Kaiming_Resnet_CVPR2016} as the structure of fPAD models $\mathcal{F}^{i} (i=1,2,3,...,K)$. In the training phase, Adam optimizer~\cite{adam} is used for the optimization of federated learning. The learning rate of federated learning is set as 1e-2. The batch size is 64 per data center. Local optimization epoch $L$ is set equal to 3. In the testing phase, Adam optimizer~\cite{adam} is used for the optimization of test time adaptation. The learning rate of test time adaptation is set as 5e-3.

\subsection{Experimental Results}

\begin{table*}[htb]
	\renewcommand{\arraystretch}{1}
	\centering	
	\caption{Comparison with models trained by data from single data center and various data centers.}
	\begin{tabular}{c|c|c|c|c|c|c|c|c}
		\hline
		\textbf{Methods}                  & \textbf{Data Centers} & \textbf{User} & \textbf{HTER (\%)} & \textbf{EER (\%)} & \textbf{AUC (\%)} & \textbf{Avg. HTER}          & \textbf{Avg. EER}          & \textbf{Avg. AUC}           \\ \hline
		\multirow{12}{*}{\textbf{Single}} & O                     & M             & 41.29              & 37.42             & 67.93             & \multirow{12}{*}{41.61} & \multirow{12}{*}{36.66} & \multirow{12}{*}{67.07} \\
		& C                     & M             & 27.09              & 24.69             & 82.91             &                         &                         &                         \\ 
		& I                     & M             & 49.05              & 20.04             & 85.89             &                         &                         &                         \\ 
		& O                     & C             & 31.33              & 34.73             & 73.19             &                         &                         &                         \\ 
		& M                     & C             & 39.80              & 40.67             & 66.58             &                         &                         &                         \\ 
		& I                     & C             & 49.25              & 47.11             & 55.41             &                         &                         &                         \\ 
		& O                     & I             & 42.21              & 43.05             & 54.16             &                         &                         &                         \\ 
		& C                     & I             & 45.99              & 48.55             & 51.24             &                         &                         &                         \\ 
		& M                     & I             & 48.50              & 33.70             & 66.29             &                         &                         &                         \\ 
		& M                     & O             & 29.80              & 24.12             & 84.86             &                         &                         &                         \\ 
		& C                     & O             & 33.97              & 21.24             & 84.33             &                         &                         &                         \\ 
		& I                     & O             & 46.95              & 35.16             & 71.58             &                         &                         &                         \\ \hline
		\multirow{4}{*}{\textbf{Fused}}    & O\&C\&I               & M             & 34.42              & 23.26             & 81.67             & \multirow{4}{*}{35.75}  & \multirow{4}{*}{31.29}  & \multirow{4}{*}{73.89}  \\ 
		& O\&M\&I               & C             & 38.32              & 38.31             & 67.93            &                         &                         &                         \\ 
		& O\&C\&M               & I             & 42.21              & 41.36             & 59.72             &                         &                         &                         \\ 
		& I\&C\&M               & O             & 28.04              & 22.24             & 86.24             &                         &                         &                         \\ \hline
		\multirow{4}{*}{\textbf{FedPAD}}    & O\&C\&I               & M             & 19.45              & 17.43             & 90.24             & \multirow{4}{*}{32.17}  & \multirow{4}{*}{28.84}  & \multirow{4}{*}{76.51}  \\ 
		& O\&M\&I               & C             & 42.27              & 36.95             & 70.49             &                         &                         &                         \\ 
		& O\&C\&M               & I             & 32.53              & 26.54             & 73.58             &                         &                         &                         \\ 
		& I\&C\&M               & O             & 34.44              & 34.45             & 71.74             &                         &                         &                         \\ \hline
		\multirow{4}{*}{\begin{tabular}[c]{@{}c@{}}\textbf{All}\\ \end{tabular}}     & O\&C\&I               & M             & 21.80              & 17.18             & 90.96             & \multirow{4}{*}{27.26}  & \multirow{4}{*}{25.09}  & \multirow{4}{*}{80.42}  \\ 
		& O\&M\&I               & C             & 29.46              & 31.54             & 76.29             &                         &                         &                         \\ 
		& O\&C\&M               & I             & 30.57              & 25.71             & 72.21             &                         &                         &                         \\ 
		& I\&C\&M               & O             & 27.22              & 25.91             & 82.21             &                         &                         &                         \\ \hline
		\multirow{4}{*}{\textbf{Ours}}    & O\&C\&I               & M             & 14.70              & 16.64             & 90.57             & \multirow{4}{*}{\textbf{23.18}}  & \multirow{4}{*}{\textbf{23.88}}  & \multirow{4}{*}{\textbf{83.40}}  \\ 
		& O\&M\&I               & C             & 26.33              & 29.75             & 77.77             &                         &                         &                         \\ 
		& O\&C\&M               & I             & 28.61              & 26.04             & 82.07             &                         &                         &                         \\ 
		& I\&C\&M               & O             & 23.09              & 23.09             & 83.21             &                         &                         &                         \\ \hline
	\end{tabular}
	\label{tab:singleallours}
\end{table*}

\begin{table*}[htb]
	\renewcommand{\arraystretch}{1}
	\centering	
	\caption{Comparison with models trained by data from single data center and various data centers.}
	\begin{tabular}{c|c|c|c|c|c|c|c|c}
		\hline
		\textbf{Methods}                                                                                  & \textbf{Data Centers} & \textbf{User} & \textbf{HTER (\%)} & \textbf{EER (\%)} & \textbf{AUC (\%)} & \textbf{Avg. HTER}              & \textbf{Avg. EER}               & \textbf{Avg. AUC}               \\ \hline
		\multirow{12}{*}{\textbf{\begin{tabular}[c]{@{}c@{}}Single\\ +Test-Time-Adaptation\end{tabular}}} & O                     & M             & 28.81              & 31.35             & 74.77             & \multirow{12}{*}{35.09}         & \multirow{12}{*}{36.43}         & \multirow{12}{*}{68.00}         \\
		& C                     & M             & 34.49              & 35.64             & 69.46             &                                 &                                 &                                 \\
		& I                     & M             & 12.11              & 16.53             & 90.23             &                                 &                                 &                                 \\
		& O                     & C             & 30.37              & 30.35             & 74.69             &                                 &                                 &                                 \\
		& M                     & C             & 41.20              & 42.10             & 60.69             &                                 &                                 &                                 \\
		& I                     & C             & 43.53              & 42.91             & 59.21             &                                 &                                 &                                 \\
		& O                     & I             & 47.88              & 46.76             & 56.71             &                                 &                                 &                                 \\
		& C                     & I             & 60.02              & 65.26             & 36.34             &                                 &                                 &                                 \\
		& M                     & I             & 17.40              & 17.04             & 89.65             &                                 &                                 &                                 \\
		& M                     & O             & 23.24              & 23.30             & 83.65             &                                 &                                 &                                 \\
		& C                     & O             & 31.63              & 31.08             & 74.34             &                                 &                                 &                                 \\
		& I                     & O             & 36.94              & 37.16             & 66.60             &                                 &                                 &                                 \\ \hline
		\multirow{4}{*}{\textbf{Ours}}                                                                    & O\&C\&I               & M             & 14.70              & 16.64             & 90.57             & \multirow{4}{*}{\textbf{23.18}} & \multirow{4}{*}{\textbf{23.88}} & \multirow{4}{*}{\textbf{83.40}} \\
		& O\&M\&I               & C             & 26.33              & 29.75             & 77.77             &                                 &                                 &                                 \\
		& O\&C\&M               & I             & 28.61              & 26.04             & 82.07             &                                 &                                 &                                 \\
		& I\&C\&M               & O             & 23.09              & 23.09             & 83.21             &                                 &                                 &                                 \\ \hline
	\end{tabular}
	\label{tab:singleTTT}
\end{table*}

\begin{figure*}[htb]
	\centering
	\begin{minipage}[t]{0.24\linewidth}
		\centering
		\includegraphics[width=1\linewidth]{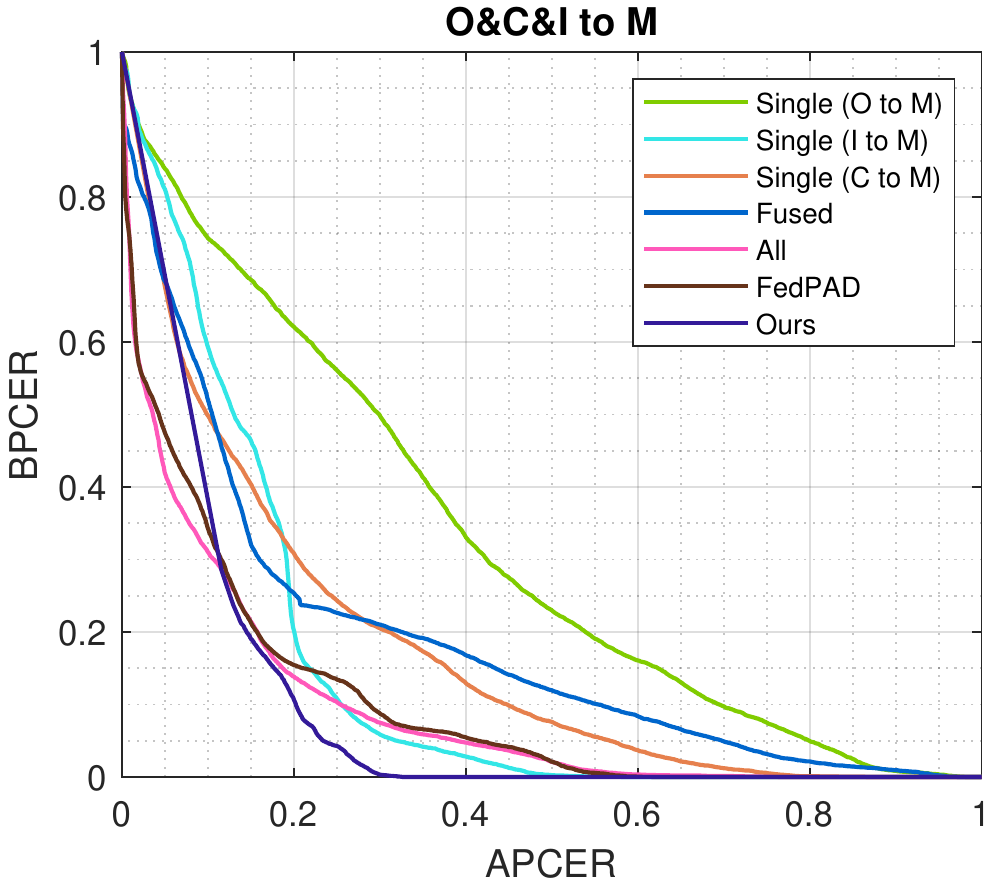}
	\end{minipage}%
	\begin{minipage}[t]{0.24\linewidth}
		\centering
		\includegraphics[width=1\linewidth]{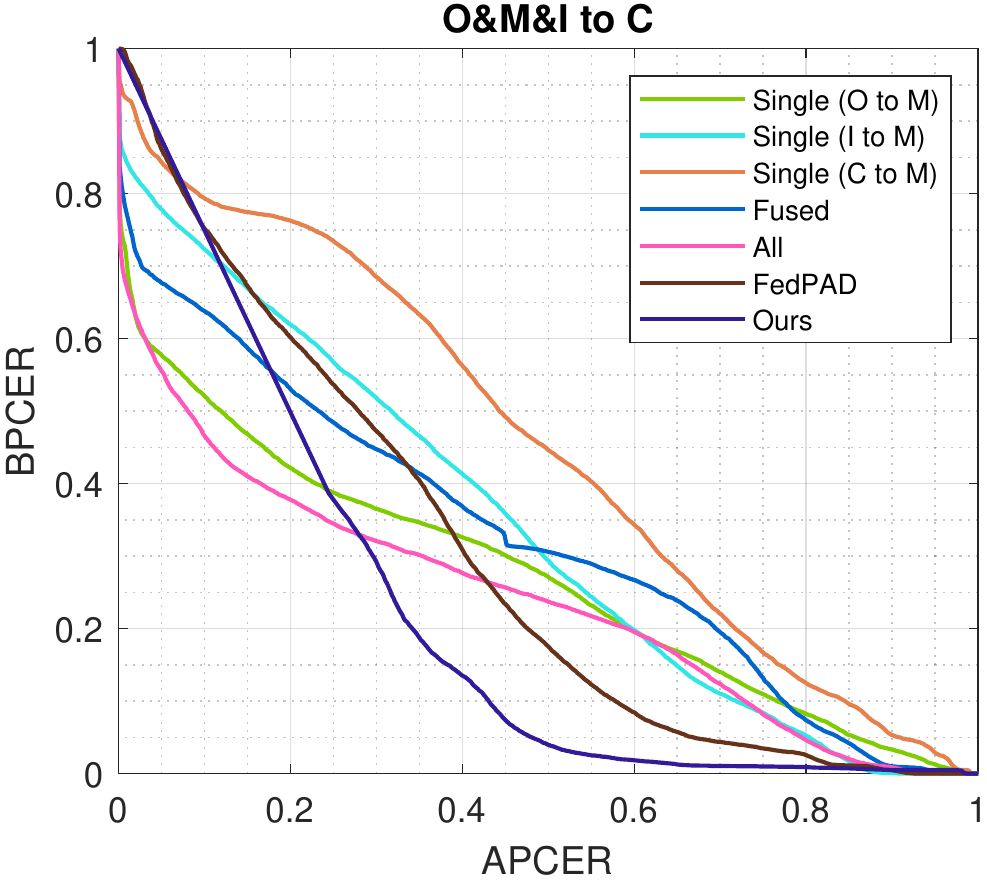}  
	\end{minipage}
	\begin{minipage}[t]{0.24\linewidth}
		\centering
		\includegraphics[width=1\linewidth]{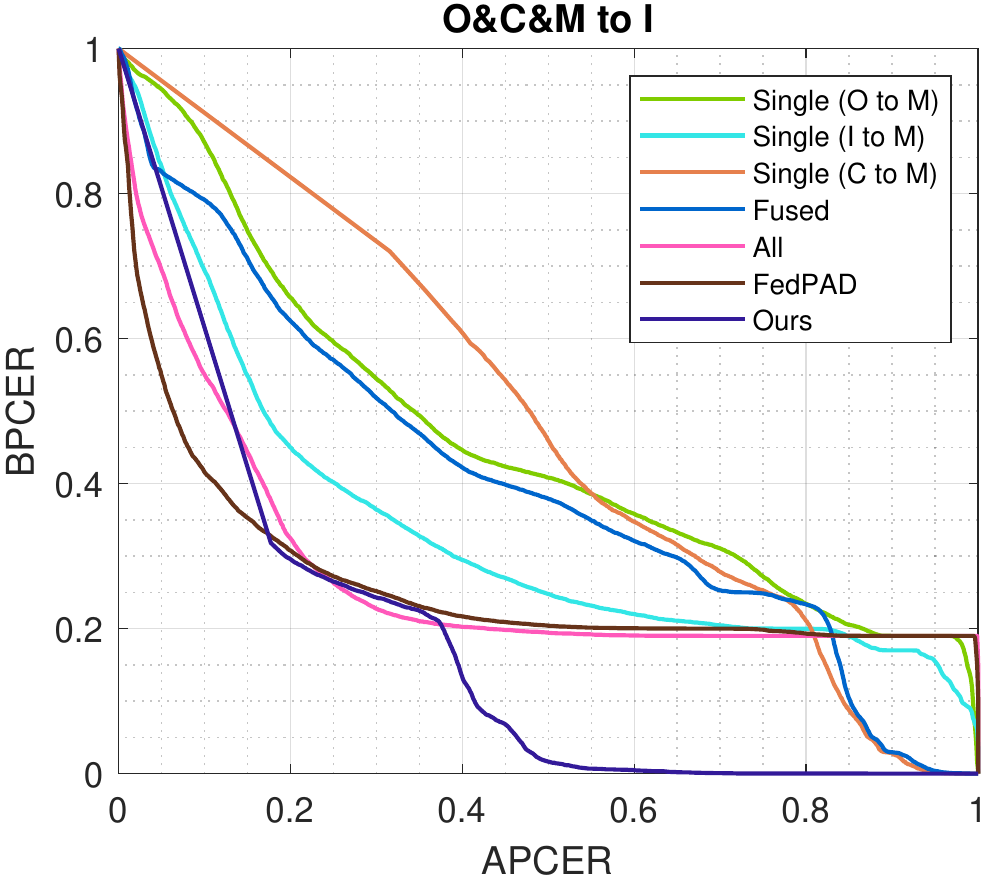}
	\end{minipage}%
	\begin{minipage}[t]{0.24\linewidth}
		\centering
		\includegraphics[width=1\linewidth]{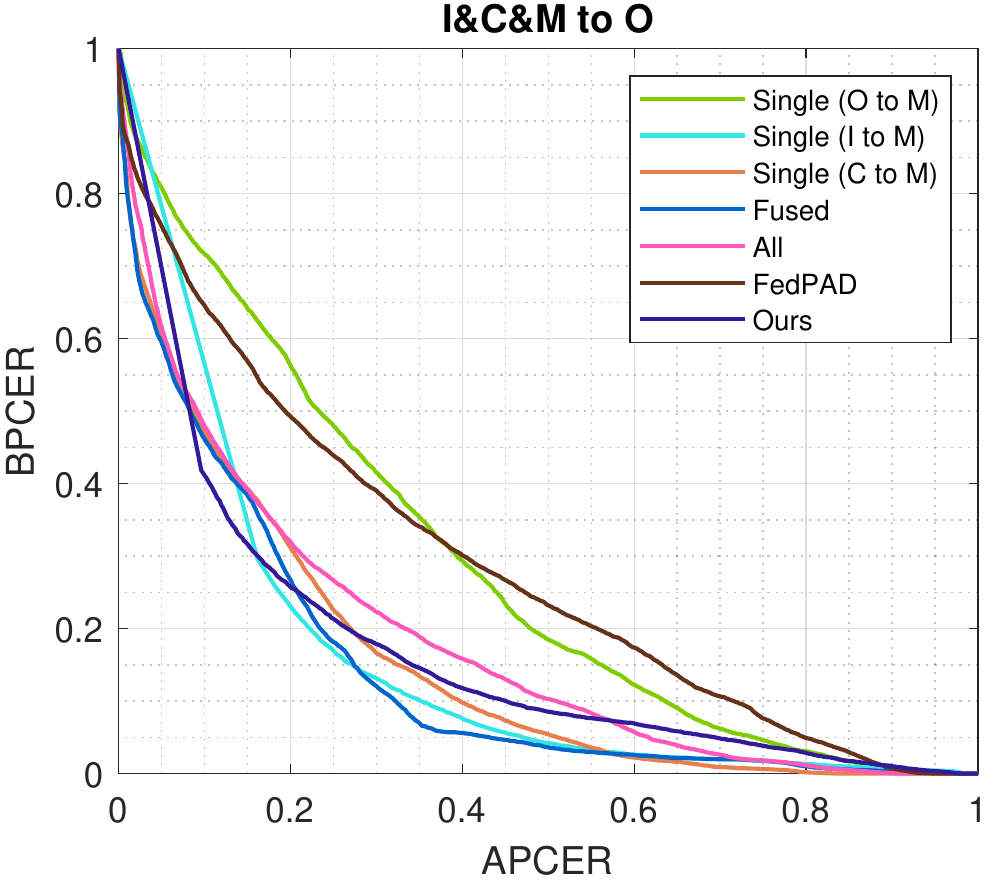}  
	\end{minipage}
	\vskip-10pt\caption{ROC curves of models trained by data from single data center and various data centers.}
	\label{fig:rocmain}
\end{figure*}


\begin{table}[htb]
	\renewcommand{\arraystretch}{1}
	\centering	
	\caption{Comparison of different number of data centers}
	\begin{tabular}{c|c|c|c|c|c}
		\hline
		\textbf{Methods}                  & \textbf{Data Centers} & \textbf{User}      & \textbf{HTER} & \textbf{EER} & \textbf{AUC} \\ \hline
		\multirow{3}{*}{\textbf{FedPAD}}  & O\&I                  & \multirow{6}{*}{C} & 49.22              & 49.10             & 51.19             \\
		& O\&M\&I               &                    & 42.27              & 36.95             & 70.49             \\
		& O\&M\&I\&S            &                    & 41.74              & 29.90             & 78.47             \\ \cline{1-2} \cline{4-6} 
		\multirow{3}{*}{\textbf{Ours}} & O\&I                  &                    & 43.09              & 42.57             & 63.21             \\
		& O\&M\&I               &                    & 26.33             & 29.75             & 77.77             \\
		& O\&M\&I\&S            &                    & \textbf{22.90}              & \textbf{22.39}             & \textbf{85.89}             \\ \hline
	\end{tabular}
	\label{tab:datacenters1}
\end{table}

\begin{table}[htb]
	\renewcommand{\arraystretch}{1}
	\centering	
	\caption{Comparison of different number of data centers.}
	\begin{tabular}{c|c|c|c|c|c}
		\hline
		\textbf{Methods}                  & \textbf{Data Centers} & \textbf{User}      & \textbf{HTER} & \textbf{EER} & \textbf{AUC} \\ \hline
		\multirow{3}{*}{\textbf{FedPAD}}  & C\&M                  & \multirow{6}{*}{S} & 25.11              & 20.64             & 88.08             \\
		& I\&C\&M               &                    & 29.61              & 14.61             & 93.30             \\
		& I\&C\&M\&O            &                    & 12.45              & \textbf{8.98}     & \textbf{97.18}    \\ \cline{1-2} \cline{4-6} 
		\multirow{3}{*}{\textbf{Ours}} & C\&M                  &                    & 15.45              & 14.06             & 92.30             \\
		& I\&C\&M               &                    & 16.45              & 15.34            & 92.03            \\
		& I\&C\&M\&O            &                    & \textbf{10.62}      & 9.63              & 96.35             \\ \hline
	\end{tabular}
	\label{tab:datacenters2}
\end{table}

In this section we demonstrate the practicality and generalization ability of the proposed framework in the real-world scenario. We first compare the performance of the proposed framework with models trained with data from a single data center. As mentioned above, due to the limitation of data privacy that exists in the real-world, data cannot be shared among different \textit{data centers}. In this case, \textit{users} will directly obtain a trained model from one of the \textit{data centers}. We report the performance of this baseline in the Table~\ref{tab:singleallours} under the label \textbf{Single}. For different choices of user datasets (from O, C, I, M), we report the performance when the model is trained from the remaining datasets independently. 

Rather than obtaining a trained model from a single data center, it is possible for users to obtain multiple trained models from several data centers and fuse their prediction scores during inference, which is also privacy preserving. In this case, we fuse the prediction scores of the trained model from various data centers by calculating the average. The results of this baseline are shown in Table~\ref{tab:singleallours} denoted as \textbf{Fused}. Please note that it is impossible for users to carry out feature fusion because a classifier cannot be trained based on the fused features without accessing to any real/spoof data during inference in the users. According to  Table~\ref{tab:singleallours}, fusing scores obtained from different data centers improves the fPAD performance on average. However, this would require higher inference time and computation complexity (of order three for the case considered in this experiment).

\begin{table*}
	\renewcommand{\arraystretch}{1}
	\centering	
	\caption{Effect of using different types of spoof attacks}
	\begin{tabular}{c|c|c|c|c|c}
		\hline
		\textbf{Methods}                 & \textbf{Data Centers}  & \textbf{User}    & \textbf{HTER (\%)} & \textbf{EER (\%)} & \textbf{AUC (\%)} \\ \hline
		\multirow{2}{*}{\textbf{Single}} & I (Print)              & M (Print, Video) & 38.82              & 33.63             & 72.46             \\ 
		& O (Video)              & M (Print, Video) & 35.76              & 28.55             & 78.86             \\ \hline
		\textbf{Fused}                     & I (Print) \& O (video) & M (Print, Video) & 35.22              & 25.56            & 81.54             \\ \hline	
		\textbf{FedPAD}                    & I (Print) \& O (video) & M (Print, Video) & 30.51    & 26.10    & 84.82   \\ \hline	
		\textbf{Ours}                    & I (Print) \& O (video) & M (Print, Video) & \textbf{18.20}     & \textbf{18.63 }   & \textbf{87.57}    \\ \hline

	\end{tabular}
	\label{tab:2Dtype}
\end{table*}

\textbf{FedPAD}~\cite{shao2020federated} exploits the federated learning on the task of fPAD. Table~\ref{tab:singleallours} illustrates that the FedPAD improves the performance on most of settings. However, when encountering large domain gap settings such as O$\&$C$\&$M to I, FedPAD still cannot achieve very promising performance. Comparatively, after integrating test-time adaptation with federated learning, on average the proposed framework (\textbf{Ours}) is able to significantly improve the performance in all settings and outperform other baselines. Moreover, we plot the ROC curves in Fig~\ref{fig:rocmain}, which also shows  the effectiveness of the proposed framework. These results demonstrate that the proposed method is more effective in facilitating the generalization ability of fPAD with the dual-phase privacy preservation.

\begin{table}[!htb]
\renewcommand{\arraystretch}{1}
\centering	
\caption{Impact of adding data centers with diverse attacks}	
	\begin{tabular}{c|c|c|c|c}
		\hline
		\textbf{Methods}                 & \textbf{Data Centers} & \textbf{User}           & \textbf{HTER} & \textbf{AUC} \\ \hline
		\multirow{2}{*}{\textbf{FedPAD}} & O\&C\&M (2D)          & \multirow{3}{*}{3 (3D)} & 27.21              & 76.05             \\
		& O\&C\&M (2D) \&H (3D) &                         & 34.70              & \textbf{92.35}    \\ \cline{1-2} \cline{4-5} 
		\textbf{Ours}                    & O\&C\&M (2D) \&H (3D) &                         & \textbf{16.97}     & 90.25             \\ \hline
	\end{tabular}
	\label{tab:3Dtype}
\end{table}

Moreover, we further consider the case where a model is trained with data from all available data centers, which is denoted as \textbf{All} in Table~\ref{tab:singleallours}. Note that this baseline violates the assumption of preserving data privacy, and therefore is not a valid comparison for FL for fPAD. Nevertheless, it indicates the upper bound of performance for the federated learning applied in fPAD. From Table~\ref{tab:singleallours}, it can be seen that the proposed framework is able to perform better than this baseline.  This shows the proposed framework is able to obtain a privacy persevering fPAD model without sacrificing fPAD performance.

\subsubsection{Compatibility of Dual Phases}
To improve the generalization ability of fPAD models, users can also download models trained with data from a single data center and carry out test-time adaptation. Therefore, a natural choice is to integrate test-time adaptation with models trained with data from a single data center (\textbf{Single}). We tabulate the comparison between this baseline (\textbf{Single+Test Time Adaptation}) and the proposed framework in Table~\ref{tab:singleTTT}. Table~\ref{tab:singleTTT} shows that the proposed framework integrating test-time adaptation with federated learning performs better than \textbf{Single+Test Time Adaptation}. fPAD models trained with data from a single data center (\textbf{Single}) will easily overfit to data of corresponding local data center and thus generate a poor fPAD model. Comparatively, by exploiting various fPAD data available from multiple data centers, federated learning can produce a more suitable fPAD model during the training phase with privacy preservation. Since test-time adaptation is very sensitive to the quality of pre-trained model, fPAD model trained by the proposed method is equipped with better initial parameters as a better starting point for the following test-time adaptation. In this way, the corresponding test-time adaptation can achieve improved performance as shown in Table~\ref{tab:singleTTT}. This demonstrates the proposed framework is more able to exploit the compatibility between dual phases and thus achieve better fAPD performance.

\subsubsection{Comparison of different number of data centers}

\begin{figure}[!htb]
	\begin{center}
		\includegraphics[ width=0.6\linewidth]{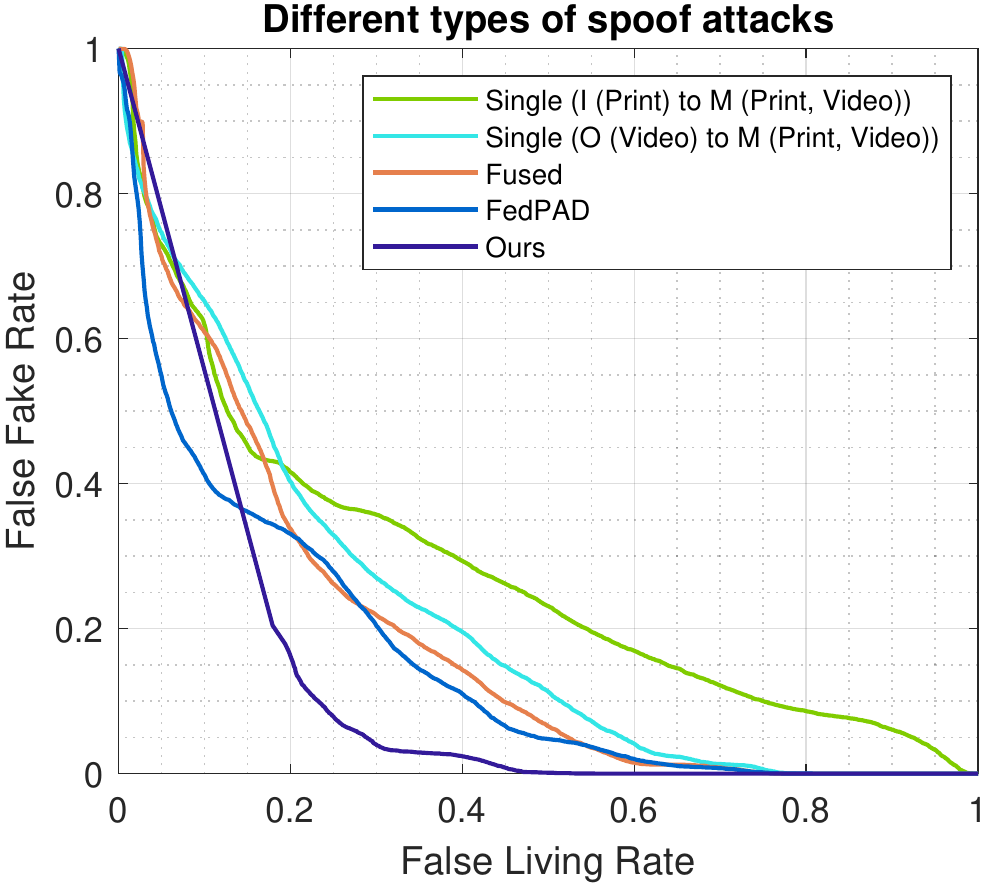}
	\end{center}
	\vskip-15pt\caption{ROC curves of models trained by different types of 2D spoof attacks.}
	\label{fig:rocprintvideo}
\end{figure}

In this section, we investigate the importance of having more data centers during training. Different data centers exploit different characteristics of face presentation attacks. Therefore, we expect aggregating information from more data centers in the proposed framework to produce more robust models with better generalization. In order to verify this point, we increase the number of data centers in the proposed framework and FedPAD method. The comparison is reported in Tables~\ref{tab:datacenters1} and~\ref{tab:datacenters2}. The experiments are carried out using five datasets (O, M, I, C, S). In Table~\ref{tab:datacenters1}, we select the dataset C as the data presented to the user and the remaining datasets as the data centers for training the fPAD model with our framework, where we increase the number of data centers from 2 to 4. Another experiment is carried out with a different combination of the same five datasets and the results are shown in Table~\ref{tab:datacenters2}. From Tables~\ref{tab:datacenters1} and~\ref{tab:datacenters2}, it can be seen that most values of evaluation metrics improve along when the number of data centers increases, and the proposed method achieves the best performance in the first setting and comparable results in the second setting compared to FedPAD. This demonstrates that increasing the number of data centers in the proposed framework can improve the performance.

\subsubsection{Generalization ability to various 2D spoof attacks}

In reality, due to limited resources, one data center may only be able to collect limited types of 2D attacks. However, various 2D attacks may appear to the users. This section supposes that one data center collects one particular type of 2D attack such as print attack or video-replay attack. As illustrated in Table~\ref{tab:2Dtype}, first, we select real faces and print attacks from dataset I and real faces and video-replay attacks from dataset O to train a fPAD model respectively and evaluate them on dataset M (containing both print attacks and video-replay attacks). In both considered cases as shown in Table~\ref{tab:2Dtype}, the corresponding trained models cannot generalize well to dataset M which contains the additional types of 2D attacks compared to dataset I and O, respectively. This tendency can be alleviated when the prediction scores of two independently trained models on both types of attacks are fused as shown in Table~\ref{tab:2Dtype}. FedPAD method obtains a performance gain compared to score fusion. Comparatively, the proposed method further improves the performance, especially with a gain of $12.31\%$ in HTER and $7.47\%$ in EER compared to FedPAD. We also plot the corresponding ROC curve for this comparison in Fig~\ref{fig:rocprintvideo}, which also demonstrate the superior performance of the proposed method. This demonstrate that test-time adaptation can effectively improve the generalization ability to various 2D attacks when FL model is trained with limited types of fPAD data.

\subsubsection{Generalization ability to 3D mask attacks }

In this section, we investigate the generalization ability of the proposed framework to 3D mask attacks and the comparison is also conducted between the proposed framework and FedPAD method. First, in FedPAD, a fPAD model is trained with data centers exploiting 2D attacks (data from datasets O, C, I and M). This model is tested with 3D mask attacks (data from dataset H). Then, we include one more data center containing 3D mask attacks (dataset 3) into FedPAD and retrain it. Table~\ref{tab:3Dtype} shows that introducing diversity of data centers  (by including a 3D mask attack) can improve performance in all evaluation metrics. This demonstrates that increasing data centers with 3D mask attacks within the federated learning framework can improve the generalization ability of fPAD model to the novel 3D mask attacks. We carry out the same experiment based on the proposed framework. In Table~\ref{tab:3Dtype}, it can be seen that the proposed method can significantly improve HTER  by $17.73\%$ compared to FedPAD. This means that after adapted with novel 3D mask attack data by test-time adaptation during testing,  fPAD model trained with federated learning in the training phase is more able to generalize well to the novel types of 3D mask attacks, which forms a better generalized fPAD framework.

\section{Conclusion}
In this paper, we presented a framework based on the principles of FL and test-time adaptation, targeting the application of fPAD with the objective of obtaining generalized fPAD models while preserving data privacy in both training and testing phases. In the training phase, through communications between \textit{data centers} and the \textit{server}, a global fPAD model is obtained by iteratively aggregating the model updates from various \textit{data centers}. In the testing phase, test-time adaptation is further exploited to minimize the prediction entropy of trained fPAD so that the generalization error on the unseen face presentation attacks can be further reduced. Local private data in the data centers is not accessed during the whole process. Extensive experiments are carried out to demonstrate the effectiveness of the proposed framework. In the future, we will further investigate the generalization improvement among different imbalanced datasets, like FedMix~\cite{yoon2021fedmix} and Fed-Focal Loss~\cite{sarkar2020fed}. Moreover, other approaches boosting the generalization ability, such as data augmentation~\cite{hendrycks2019augmix} and robust loss functions~\cite{zhang2018generalized}, will also be studied in the future work.

{\small
\bibliographystyle{ieee}
\bibliography{egbib}

\begin{thebibliography}{10}\itemsep=-1pt

\bibitem{2004HTER}
S.~Bengio and J.~Mari{\'{e}}thoz.
\newblock A statistical significance test for person authentication.
\newblock In {\em The Speaker and Language Recognition Workshop}, 2004.

\bibitem{2017FGoulu}
Z.~Boulkenafet and et~al.
\newblock Oulu-npu: A mobile face presentation attack database with real-world
  variations.
\newblock In {\em FG}, 2017.

\bibitem{2016TIFScolortxt}
Z.~Boulkenafet, J.~Komulainen, and A.~Hadid.
\newblock Face spoofing detection using colour texture analysis.
\newblock In {\em IEEE Trans. Inf. Forens. Security, 11(8): 1818-1830}, 2016.

\bibitem{2012BIOSIGidiap}
I.~Chingovska, A.~Anjos, and S.~Marcel.
\newblock On the effectiveness of local binary patterns in face anti-spoofing.
\newblock In {\em BIOSIG}, 2012.

\bibitem{Marceltifs3D2014}
N.~Erdogmus and S.~Marcel.
\newblock Spoofing face recognition with {3D} masks.
\newblock 2014.
\newblock TIFS.

\bibitem{Kaiming_Resnet_CVPR2016}
K.~He, X.~Zhang, S.~Ren, and J.~Sun.
\newblock Deep residual learning for image recognition.
\newblock In {\em CVPR}, 2016.

\bibitem{hendrycks2019augmix}
D.~Hendrycks, N.~Mu, E.~D. Cubuk, B.~Zoph, J.~Gilmer, and B.~Lakshminarayanan.
\newblock Augmix: A simple data processing method to improve robustness and
  uncertainty.
\newblock {\em arXiv preprint arXiv:1912.02781}, 2019.

\bibitem{adam}
D.~P. Kingma and J.~Ba.
\newblock {A}dam: A method for stochastic optimization.
\newblock In {\em arXiv preprint arXiv:1412.6980}, 2014.

\bibitem{li2019federated}
T.~Li, A.~K. Sahu, A.~Talwalkar, and V.~Smith.
\newblock Federated learning: Challenges, methods, and future directions.
\newblock In {\em arXiv preprint arXiv:1908.07873}, 2019.

\bibitem{2018ECCVrPPG}
S.~Liu, X.~Lan, and P.~C. Yuen.
\newblock Remote photoplethysmography correspondence feature for {3D} mask face
  presentation attack detection.
\newblock In {\em ECCV}, 2018.

\bibitem{2016ECCVrPPG}
S.~Liu, P.~C. Yuen, S.~Zhang, and G.~Zhao.
\newblock 3{D} mask face anti-spoofing with remote photoplethysmography.
\newblock In {\em ECCV}, 2016.

\bibitem{2018CVPRauxliary}
Y.~Liu, A.~Jourabloo, and X.~Liu.
\newblock Learning deep models for face anti-spoofing: Binary or auxiliary
  supervision.
\newblock In {\em CVPR}, 2018.

\bibitem{liu2019deep}
Y.~Liu, J.~Stehouwer, A.~Jourabloo, and X.~Liu.
\newblock Deep tree learning for zero-shot face anti-spoofing.
\newblock In {\em CVPR}, 2019.

\bibitem{2011IJCBmstexture}
J.~M{\"{a}}{\"{a}}tt{\"{a}}, A.~Hadid, and M.~Pietik{\"{a}}inen.
\newblock Face spoofing detection from single images using micro-texture
  analysis.
\newblock In {\em IJCB}, 2011.

\bibitem{mcmahan2016communication}
H.~B. McMahan, E.~Moore, D.~Ramage, S.~Hampson, et~al.
\newblock Communication-efficient learning of deep networks from decentralized
  data.
\newblock In {\em AISTATS}, 2016.

\bibitem{mohri2019agnostic}
M.~Mohri, G.~Sivek, and A.~T. Suresh.
\newblock Agnostic federated learning.
\newblock 2019.

\bibitem{2014EJIVPlbptop}
T.~F. Pereira and et~al.
\newblock Face liveness detection using dynamic texture.
\newblock In {\em EURASIP Journal on Image and Video Processing, (1): 1-15},
  2014.

\bibitem{sahu2018convergence}
A.~K. Sahu, T.~Li, M.~Sanjabi, M.~Zaheer, A.~Talwalkar, and V.~Smith.
\newblock On the convergence of federated optimization in heterogeneous
  networks.
\newblock In {\em arXiv preprint arXiv:1812.06127}, 2018.

\bibitem{sarkar2020fed}
D.~Sarkar, A.~Narang, and S.~Rai.
\newblock Fed-focal loss for imbalanced data classification in federated
  learning.
\newblock {\em arXiv preprint arXiv:2011.06283}, 2020.

\bibitem{shannon1948mathematical}
C.~E. Shannon.
\newblock A mathematical theory of communication.
\newblock {\em The Bell system technical journal}, 27(3):379--423, 1948.

\bibitem{Shao2019CVPR}
R.~Shao, X.~Lan, J.~Li, and P.~C. Yuen.
\newblock Multi-adversarial discriminative deep domain generalization for face
  presentation attack detection.
\newblock In {\em CVPR}, 2019.

\bibitem{RuiShao2018IJCB}
R.~Shao, X.~Lan, and P.~C. Yuen.
\newblock Deep convolutional dynamic texture learning with adaptive
  channel-discriminability for 3{D} mask face anti-spoofing.
\newblock In {\em IJCB}, 2017.

\bibitem{2018TIFSdynamictext}
R.~Shao, X.~Lan, and P.~C. Yuen.
\newblock Joint discriminative learning of deep dynamic textures for {3D} mask
  face anti-spoofing.
\newblock In {\em IEEE Trans. Inf. Forens. Security, 14(4): 923-938}, 2019.

\bibitem{Shao_2020_AAAI}
R.~Shao, X.~Lan, and P.~C. Yuen.
\newblock Regularized fine-grained meta face anti-spoofing.
\newblock In {\em AAAI}, 2020.

\bibitem{shao2020federated}
R.~Shao, P.~Perera, P.~C. Yuen, and V.~M. Patel.
\newblock Federated face presentation attack detection.
\newblock {\em arXiv preprint arXiv:2005.14638}, 2020.

\bibitem{smith2017federated}
V.~Smith, C.-K. Chiang, M.~Sanjabi, and A.~S. Talwalkar.
\newblock Federated multi-task learning.
\newblock In {\em NIPS}, 2017.

\bibitem{wang2021tent}
D.~Wang, E.~Shelhamer, S.~Liu, B.~Olshausen, T.~Darrell, U.~Berkeley, and
  A.~Research.
\newblock tent: fully test-time adaptation by entropy minimization.
\newblock In {\em International Conference on Learning Representations}, 2021.

\bibitem{2015TIFSida}
D.~Wen, H.~Han, and A.~K. Jain.
\newblock Face spoof detection with image distortion analysis.
\newblock In {\em IEEE Trans. Inf. Forens. Security, 10(4): 746-761}, 2015.

\bibitem{yoon2021fedmix}
T.~Yoon, S.~Shin, S.~J. Hwang, and E.~Yang.
\newblock Fedmix: Approximation of mixup under mean augmented federated
  learning.
\newblock {\em arXiv preprint arXiv:2107.00233}, 2021.

\bibitem{2012ICBcasia}
Z.~Zhang and et~al.
\newblock A face antispoofing database with diverse attacks.
\newblock In {\em ICB}, 2012.

\bibitem{zhang2018generalized}
Z.~Zhang and M.~R. Sabuncu.
\newblock Generalized cross entropy loss for training deep neural networks with
  noisy labels.
\newblock In {\em 32nd Conference on Neural Information Processing Systems
  (NeurIPS)}, 2018.

\end{thebibliography}
}

\end{document}